\documentclass{article}

\usepackage{datetime}
\newdateformat{monthyear}{\monthname[\THEMONTH] \THEYEAR}
\newdate{date}{30}{10}{2022}
\date{\displaydate{date}}

\PassOptionsToPackage{numbers, compress}{natbib}
\usepackage[preprint]{neurips_2022}

\usepackage[utf8]{inputenc} % allow utf-8 input
\usepackage[T1]{fontenc}    % use 8-bit T1 fonts
\usepackage{hyperref}       % hyperlinks
\usepackage{url}            % simple URL typesetting
\usepackage{booktabs}       % professional-quality tables
\usepackage{amsfonts}       % blackboard math symbols
\usepackage{nicefrac}       % compact symbols for 1/2, etc.
\usepackage{microtype}      % microtypography
\usepackage{xcolor}         % colors
\usepackage{enumitem}
\usepackage{adjustbox}
\usepackage{subcaption}
\usepackage{marginnote}

\usepackage{cancel}
\usepackage{amssymb}
\usepackage{nomencl}
\usepackage{amsmath}
\usepackage{mathrsfs} % \mathscr{}
\usepackage{fixmath}  % \mathbold{}
\usepackage{bm}       % \bm{}
\usepackage{nicematrix}
\makenomenclature

\newcommand{\R}{\mathbb{R}}
\newcommand{\pr}{\mathbb{P}}
\newcommand{\Mix}{\boldsymbol{\mathrm{P}}}
\newcommand{\Cov}{\mathbold{\Sigma}}
\newcommand{\Gram}{\mathbold{G}}
\newcommand{\Kern}{\boldsymbol{\mathrm{k}}}
\newcommand{\Real}{\boldsymbol{\mathrm{Re}}}
\newcommand*{\Vline}{\rule[-1ex]{0.5pt}{2.5ex}}

\newcommand{\indep}{\perp \!\!\! \perp}
\newcommand\numberthis{\addtocounter{equation}{1}\tag{\theequation}}
\usepackage{mathtools}

\newcommand{\tocite}[1]{[{\textcolor{red}{?}}]}
\title{A Unified View of Long-Sequence Models towards Modeling Million-Scale Dependencies 
}

\author{%
  Hongyu H\`e \thanks{Correspondence to \texttt{<honghe@inf.ethz.ch>}}\\
  Department of Computer Science\\
    {ETH Zurich}\\
  \And Marko Kabic\\
  Department of Computer Science\\
    {ETH Zurich}\\
}

\begin{document}
\maketitle
% {\centerline{\monthyear\today}}

\nomenclature{$\vec{a}_{i}$}{Element at $i$th position of column vector $\vec{a}$}
\nomenclature{$F^h_{D\times D}$}{Vandermonde matrix of the embedding dimension}
\nomenclature{$F^s_{L\times L}$}{Vandermonde matrix of the sequence dimension}

\nomenclature{$\Cov$}{Covariance matrix}
\nomenclature{$\Gram$}{Gram/kernel matrix}
\nomenclature{$\Mix(\cdot)$}{Token mixing process}
\nomenclature{$\Kern(\cdot)$}{Kernel function}
\nomenclature{$\Real(\cdot)$}{Function that extracts the real component of a complex number}
\nomenclature{$\texttt{FT}(\cdot)$}{(Discrete) Foureir Tansform}
\nomenclature{$\mathbb{P}(\cdot)$}{Probability density}

\nomenclature[a]{$A_{N\times M}$}{Shorthand for matrix $A \in \R^{N\times M}$}
\nomenclature[a]{$A_{i, j}$}{Element in $i$th row $j$th column of $A$}
\nomenclature[a]{$A_{i}$}{Row vector in $i$th row of $A$ ($= A_{i: *}$)}
\nomenclature[a]{$A_{*:j}$}{Column vector in $j$th row of $A$}

\nomenclature[x]{$X_{L\times D}$}{\ Input sequence of length $L$ and embedding dimension $D$, where $L \gg D$}
\nomenclature[x]{$\widetilde{X}$}{\ Resulting tokens with inductive bias introduced into $X$}

\nomenclature[w]{$W$}{Weight matix learned with element-wise non-linearity (e.g., ReLU, GELU)}
\nomenclature[w]{$W^{\mathcal{C}}_{L\times L}$}{Weight matix of a single convolution kernel}
\nomenclature[w]{$W^{\mathcal{V}}_{D\times M}$}{Weight matix of attention value}
\nomenclature[w]{$W^{\mathcal{Q}}_{D\times M}$}{Weight matix of attention query}
\nomenclature[w]{$W^{\mathcal{K}}_{D\times N}$}{Weight matix of attention key (for self-attention, $N=M$)}
% \nomenclature{$h$}{Planck constant}

\vspace{1cm}
\printnomenclature
\clearpage

\begin{abstract}
    Ever since their conception, Transformers have taken over traditional sequence models in many tasks, such as NLP, image classification, and video/audio processing, for their fast training and superior performance.
    Much of the merit is attributable to positional encoding and multi-head attention.
    However, Transformers fall short in learning long-range dependencies mainly due to the quadratic complexity scaled with context length, in terms of both time and space.
    Consequently, over the past five years, a myriad of methods has been proposed to make Transformers more efficient.
    In this work, we first take a step back, study and compare existing solutions to long-sequence modeling in terms of their pure mathematical formulation.
    Specifically, we summarize them using a unified template, given their shared nature of token mixing.
    Through benchmarks, we then demonstrate that long context length does yield better performance, albeit application-dependent, and traditional Transformer models fall short in taking advantage of long-range dependencies.
    Next, inspired by emerging sparse models of huge capacity, we propose a machine learning system for handling million-scale dependencies.
    As a proof of concept, we evaluate the performance of one essential component of this system, namely, the distributed multi-head attention.
    We show that our algorithm can scale up attention computation by almost $40\times$ using four GeForce RTX 4090 GPUs, compared to vanilla multi-head attention mechanism.
    We believe this study is an instrumental step towards modeling million-scale dependencies.
\end{abstract}

\section{Introduction}

Sequence models (e.g., LSTM \cite{hochreiter1997lstm}, GRU \cite{chung2014gru}) can capture relationships among input tokens during the learning process, where the tokens can be words, pixels, signals, etc.
By utilizing the context information, sequence models learn complex dependencies in training samples, which are referenced during inference.
Such in-context learning has demonstrated superior performance in many tasks (e.g., NLP \cite{huang2015bidirectional}, image processing \cite{venugopalan2015seqimg}).
However, traditional sequence models have two main drawbacks, namely, slow training and forgetting long-range dependencies.
The former is due to the sequential nature of those models --- tokens must be fed sequentially and in order.
The latter is the result of their limited model capacity (e.g., that of the hidden state in LSTM).

Ever since its conception, Transformers \cite{vaswani2017attention} have quickly taken over traditional sequence models for their fast training and superior performance.
The speedup was due to the use of positional encoding \cite{vaswani2017attention,xu2021positional}. 
This technique allows for parallel processing of input tokens without losing their information related to their semantic ordering, by encoding token positions directly into the token embedding.
Positional encodings slightly nudge tokens in the feature space towards a direction based on their positions in a sequence, without destroying the information encoded in their original embedding vector space.
It can also be applied to structured inputs \cite{ainslie2020etc} like images by using relative positional information \cite{wu2021rethinkingpositional}, similar to convolutional kernels.
Furthermore, their great performance can be attributed to the attention mechanism, or more specifically, multi-head self/cross-attention.
Attention mechanism explicitly learns the dependencies in input by computing pairwise attention scores among for all combinations of tokens.
The attention scores can be computed in many ways \cite{tsai2019kernattn} and the most straightforward, yet very effective, method is the dot product, which measures the distance between two embedding vectors.
This mechanism greatly improves the performance of sequence modeling and is necessary for certain tasks to be feasible, for example, audio and video processing.

Despite their great advantages, Transformers have critical disadvantages as well, and chief among them is their resource efficiency.
Although Transformers are much faster compared to traditional sequence models, the attention matrix incurs a $O(L^2)$ complexity, where $L$ is the context length, in terms of both compute and storage.
Such a complexity is especially prohibitive when dealing with context of thousands of tokens since the time and memory it takes scale quadratically, for example, summarizing books, processing audio/video, dealing with high-resolution images, etc.
In these tasks, the context can easily scale to thousands or even millions of tokens as making a decision may need information from far earlier steps in the sequence (e.g., a name in the first chapter of a book).
To counter this challenge, over the past five years a great deal of studies has focused on making Transformers more efficient.
The proposed methods include (but not limited to) for example, approximating the attention matrix with sparsity \cite{ho2019axial,beltagy2020longformer,zaheer2020bigbird}, clustering before computing attention \cite{roy2021efficient,kitaev2020reformer}, making assumptions via conditional probability \cite{ren2021combiner}, low-rank estimation \cite{wang2020linformer}, better memory I/O \cite{dao2022flashattention}, matrix orthogonality and associativity \cite{choromanski2020performer}, etc.
Apart from tackling the efficiency problem of Transformers directly, many other models have been proposed to cater the need for long-context learning. 
To name a few, MLP-Mixer \cite{tolstikhin2021mlpmixer}, FNet \cite{lee2021fnet} and SGConv \cite{li2022sgconv} learn from and (solely) utilize the token-mixing paradigm from Transformers; 
Memorizing Transformers \cite{wu2022memorizing} take Neural Turing Machines \cite{graves2014ntm} to extreme and make Transformers a huge LSTM-like structure;
S4 \cite{gu2022s4,gu2022s4d} rejuvenates traditional State Space Models combined with orthogonal polynomial projection to reconstruct histories.

Acknowledging the significance in both the novelty and volume of prior work, we take a step back and compile this work with the following objectives:
\begin{enumerate}%[align=left,itemindent=0pt,listparindent=0pt,labelsep=0pt,topsep=0pt,partopsep=0pt,parsep=0pt,itemsep=0pt]
    \item Categorizing existing solutions to long-range dependency problems purely by their mathematical formulations.
    \item Comparing various methods using a unified template, with which we study the tradeoffs in capturing global and local dependencies in input sequence.
    \item Empirically evaluating the impact of context length on sequence modeling in experiments with state-of-the-art models.
    \item Proposing a machine learning system for learning million-scale dependencies, aiming to strike a balance between model quality and resource efficiency.
    \item Designing and testing the feasibility of a distributed algorithm for computing the attention matrix for sequences of millions of tokens.
\end{enumerate}

\section{Problem Formulation} \label{sec:problem}

% This analysis is compiled for the following objectives:
% \begin{enumerate}%[align=left,itemindent=0pt,listparindent=0pt,labelsep=0pt,topsep=0pt,partopsep=0pt,parsep=0pt,itemsep=0pt]
%     \item Categorizing existing solutions to long-range dependency problems at a macro level.
%     \item Unifying various categories using the same template formulated below for cleaner comparisons.
%     \item With the unified template, investigating the different ways and their trade-offs in capturing global and local dependencies among input tokens.
% \end{enumerate}

In general, existing attention-like methods take a sequence $X$ of length $L$ and dimension $D$ as an input.
These methods introduce inductive bias (i.e., dependencies) to augment the original feature space of $X$ through a ``mixing'' process $\Mix$ that is either parameterized by $\theta$ learned with various techniques (Eq. \ref{eq:learned}) or fixed using certain procedures, e.g., Fourier Transform (Eq. \ref{eq:fixed}).

    \begin{align}
        \Mix(X\mid \theta): \mathcal{X} &\mapsto{} \mathcal{\widetilde{X}} \label{eq:learned} \\
        \Mix(X): \mathcal{X} &\mapsto{} \mathcal{\widetilde{X}} \label{eq:fixed} 
    \end{align}

Moreover, we further categorize $\Mix$ into the following four paradigms:

\begin{enumerate}%[align=left,itemindent=0pt,listparindent=0pt,labelsep=0pt,topsep=0pt,partopsep=0pt,parsep=0pt,itemsep=0pt]
    \item $\Mix(\cdot\mid \theta) \indep X$: Learned mixing independent of the input, e.g., simple convolution (\S\ref{sec:conv}) and MLP-Mixer (\S\ref{sec:mlp}).
    \item $\Mix(\cdot\mid \theta)\ \cancel{\indep}\ X$: Learned mixing dependent on the input, e.g., self-attention (\S\ref{sec:attn}).
    \item $\Mix(\cdot) \indep X$: Fixed mixing independent of the input, e.g., FNet (\S\ref{sec:fourier}).
    \item $\Mix(\cdot)\ \cancel{\indep}\ X$: Fixed mixing dependent on the input, e.g., State Space Model with fixed transition matrices (\S\ref{sec:ssm}).
\end{enumerate}

\section{Convolution on Signals} \label{sec:conv}

First of all, we draw parallels between convolutions and the attention mechanism.
Without the loss of generality, we assume that the input has been linearized to 1D (e.g., \citep{dosovitskiy2020vit,touvron2020deit,han2021tnt}).
For simplicity, we only consider depthwise convolution (without pointwise convolution), i.e., $X_{L\times1}$.
Given a filter/kernel $f$ of window size $K$, the representation $Y_t$ of $t$th token resulting from the convolution on signal $g$ is a weighted average over the input:

\begin{equation}
    Y_t := (f_w \odot g_X) (t) = \sum^{K-1}_{k:=0} \vec{w}_k \cdot X_{-k+t}\quad \mathrm{(depthwise)}. \label{eq:conv-depth}
\end{equation}

In addition, the following properties of convolution will be used in later derivations.
\begin{itemize}
    \item Commutativity: Eq. \ref{eq:conv-depth} can be rewritten as $Y_t = \sum^{K-1}_{k:=0} \vec{w}_{-k+t} \cdot X_k$ \citep{dai2021coatnet}.
    \item Summation distributivity: $\sum (f \odot g) \equiv \sum f \cdot \sum g$.
    \item Convolution Theorem: $f \odot g \equiv \texttt{FT}(f \cdot g)$. 
\end{itemize}

% \subsection{Convolution as Structured Matrices}

A convenient way of viewing the convolution on the entire sequence $X$ is to formulate Eq. \ref{eq:conv-depth} as weighing the input with structured weight matrices:

\begin{align}
    Z := f_W \odot g_X
        % &=   
        %     \left[
        %       \begin{array}{ccc}
        %         \Hline
        %         X_t
        %         \Hline \\
        %       \end{array}
        %     \right] 
        % \begin{bNiceMatrix} 
        %   w_1 & w_2 & \Hline & \w_k   &      & \Cdots & 0\\
        % %   0   & w_1 & w_2    & \Hline & \w_k & \Cdots & 0
        %  \end{bNiceMatrix} \nonumber
        &=                  
        \begin{bNiceMatrix} 
           w_1 & 0   & 0 & \Cdots & 0\\
           w_2 & w_1 & 0 & & \\
        \Vline & w_2 & w_1 & & \\
        \Vline &\Vline& w_2 & & \\
        \Vline & \Vline & \Vline & & \Vdots\\
           w_k &  & & \\
             0 & w_k & & & 0\\
               & 0   & w_k & & w_1\\
               &     & 0   & & w_2\\
        \Vdots & \Vdots & \Vdots & \Ddots & \Vline\\
             0 & 0 & 0 & \Cdots & w_k
         \end{bNiceMatrix} \cdot
         \left[
              \begin{array}{ccc}
                \Vline \\
                X_t \\
                \Vline \\
              \end{array}
            \right] \nonumber\\
        &= W^{\mathcal{C}}{X} \label{eq:conv-matrix}\\
        &= \Mix_\textrm{conv}(X\mid \theta_\textrm{conv}) \nonumber \\
        &= \textcolor{blue}{\widetilde{X}_{\mathrm{conv}}} , \label{eq:conv-mix}
\end{align}
where $\theta_\textrm{conv} = \{W^{\mathcal{C}}\}$.

Observe that (1) $\Mix_\textrm{conv}$ is \textit{learned} but \textit{independent} of the input sequence, since $\textcolor{blue}{X}$ does not enter $\theta_\textrm{conv}$, and (2) the window size $K$ is \textit{fixed} across all input signals.   
Consequently, first few layers of a CNN have limited \textit{local views} of the input sequence, and only by stacking kernels can enlarge the receptive fields of the higher layers.

The idea of stacking kernels to expand the receptive fields gives rise to a host of methods (e.g., \citep{wang2020linformer,beltagy2020longformer,choromanski2020performer}) to approximate the full-attention matrix, which are analogues of Eq. \ref{eq:conv-matrix}.

\section{Self-Attention} \label{sec:attn}

Similar to convolution discussed in \S\ref{sec:conv}, the attention mechanism can also be viewed as a weighted average over the raw embeddings of the input tokens.
For simplicity, masking and scaling are excluded.

\subsection{Attention as Weighted Average}

Firstly, three matrices of the same size (self-attention) are obtained by linearly projecting the \textit{same} input $X$ three times with \textit{learnable} weight matrices via MLP layers:
\begin{equation*}
    V := XW^{\mathcal{V}}, \quad
    K := XW^{\mathcal{K}}, \quad
    Q := XW^{\mathcal{Q}}
\end{equation*}

The second step is to compute the attention scores for each token in the sequence at position $t$:
\begin{equation}
    A'_t := Q_t K^\top. \label{eq:unnormalized}
\end{equation}

Then, the attention scores are normalized row-wise using \texttt{softmax}:
\begin{equation*}
    A_{t, i} := \cfrac{\exp{A'_{t, i}}}{\sum^{L-1}_{j=0} \exp{A'_{t, j}}}.
\end{equation*}

Lastly, the input projections $V$ is weighted by the score to induce biases/context, producing the final representation of token $t$:
\begin{equation*}
    Z_t := A_tV.
\end{equation*}

Putting steps together, we arrive at the weighted average similar to Eq. \ref{eq:conv-depth}:
\begin{equation}
    Z_t := \sum^{L-1}_{j=0} \underbrace{\texttt{softmax}\left\{Q_t K^\top\right\}}_{\mathrm{full-attention\ weights}} \cdot V_i. \label{eq:attn-sum}
\end{equation}

\subsection{Attention as Token Mixing}

Different from Eq. \ref{eq:conv-depth}, here the input sequence $X$ directly enters the weights in Eq. \ref{eq:attn-sum}.
For simplicity, from now on, we omit any kinds of normalization such as \texttt{softmax}, layer/batch norm, etc.

Then, we can rewrite Eq. \ref{eq:attn-sum} to:

\begin{align}
    Z_{L \times D}  :&= \underbrace{Q_t K^\top}_{A' (Eq. \ref{eq:unnormalized})} \cdot V \nonumber \\
        &= \left[(\textcolor{blue}{X}W^{\mathcal{Q}}) (\textcolor{blue}{X}W^{\mathcal{K}})^\top \right] \cdot \textcolor{blue}{X} W^{\mathcal{V}} \nonumber \\
        &= \left[{X} (W^{\mathcal{Q}} W^{\mathcal{K}\top}) X^\top\right] X  W^{\mathcal{V}} \nonumber \\
        &= \left[{X} \Gram^\mathcal{W} X^\top \right] X W^{\mathcal{V}} \label{eq:gram-weight} \\
        &= A'\ \textcolor{blue}{X}\ W^{\mathcal{V}} \label{eq:axw}\\
        &= \Mix_\textrm{attn}(X\mid \theta_\textrm{attn}) \nonumber \\
        &= \textcolor{blue}{\widetilde{X}_\mathrm{attn}} \nonumber,
\end{align}

where $\theta_\textrm{attn} = \{A', W^{\mathcal{V}}\}$, and Eq. \ref{eq:gram-weight} summarizes the product between the two learned weight matrices into their Gram matrix $\Gram^\mathcal{W}_{D \times D}$. 
% , and similarly, Eq. \ref{eq:gram-input} summarizes the kernel (that depends on the product between each pair of tokens) enclosed by square brackets into $\Gram^{\mathcal{X}}_{L \times L}$.

As a result, we obtain a new representation $\textcolor{blue}{\widetilde{X}_\mathrm{attn}}$ of the input sequence by \textit{learning a mixing scheme} over the raw tokens using self-attention (Eq. \ref{eq:axw}).
In other words, $\Mix_\textrm{attn}$ is \textit{learned} and \textit{dependent} on the input since $\textcolor{blue}{X}$ enters the equaltion via $\theta_\textrm{attn}$.

\subsection{Using Static Kernel or Feature Map} 

The computational complexity of Eq. \ref{eq:axw} is in $O(L^2D) \equiv O(L^2)$ due to the product between the (unnormalized) full-attention matrix $A'_{L\times L}$ and the input sequence $X_{L\times D}$.

Similar to Eq. \ref{eq:gram-weight}, we can \textcolor{black}{regard $A'$ as parameterized Gram matrix of the input space $\Gram^\mathcal{X}$, since it \textit{only} depends on the tokens}. 
% \textcolor{orange}{$A'$ as a feature map over the Gram matrix of the input sequence $\Gram^\mathcal{X}$}. 
In other words, the full attention could be expressible by a \textit{kernel function}. 
Specifically, we rewrite Eq. \ref{eq:gram-weight} as:
\begin{align}
    Z   &= [ \underbrace{({X} \Gram^\mathcal{W} X^\top)}_{\textcolor{black}{A':\ \mathrm{parameterized\ \Gram^\mathcal{X}}}}] X \cancel{W^{\mathcal{V}}} \nonumber\\
        % &= \left({X} \Cov^\mathcal{W} X^\top\right) X W^{\mathcal{V}} \nonumber\\
        % &= A' X W^{\mathcal{V}} \nonumber\\
        &\approxeq \Kern(X, X) \cdot X, \label{eq:kern-seq}\\ 
        &=\phi(X) \phi(X)^\top \cdot X  \label{eq:map-seq}    
\end{align}

where $\Kern$ is a kernel function over the input tokens, and $\phi$ is the equivalent feature map. For $\cancel{W^{\mathcal{V}}}$, \citet{tsai2019kernattn} have shown that this linear projection is redundant and can lead to performance degradation.
% where $\phi$ is a learned function of $\Gram^\mathcal{X}_{D\times D}$ in place of $A'_{L\times L}$ and even the projection $W^{\mathcal{V}}$.

Instead of learning three projections and computing the full-attention matrix, we could potentially learn or use a static kernel function $\Kern(\cdot)$ to capture the correlations between tokens.
Alternatively, since $L \gg D$, applying a feature map $\phi(\cdot)$ should be much cheaper than using kernel functions.
% \answerTODO{Complexity comparisons here}
 
% Can we learn a projection over $\Cov^\mathcal{X}$ as $\phi$?
% \begin{equation}
%     A' X W^{\mathcal{V}} \approx X \phi(\Cov^\mathcal{X}) 
%                          = X \bm{\mathrm{\Phi}}_{\Cov^\mathcal{X}} . \label{eq:phi}
% \end{equation}

% Since the linear map $\bm{\mathrm{\Phi}}_{\Cov^\mathcal{X}}$ or $\bm{\mathrm{\Phi}}_{\Cov^\mathcal{X}}$ is in $\R^{D\times D}$, the computational cost of Eq. \ref{eq:phi} would be in $O(D^2L) \equiv O(L)$, assuming $L >> D^2$.

\section{Sequence Modeling with Multilayer Perceptrons} \label{sec:mlp}

\citet{tolstikhin2021mlpmixer} is the first to suggest that full-attention can be replaced by \textit{learned} token mixing by only using multilayer perceptrons (MLPs), namely, MLP-Mixer (Mixer).
For simplicity, common tricks like layer norms and skip connections are omitted here.

% \subsection{Two-Step Mixing}

Mixer was initially intended for imaging tasks \citep{tolstikhin2021mlpmixer}, so the input tokens are sequentialized image patches.
However, this scheme can be generalized to any sequence-to-sequence tasks \citep{lee2021fnet}.
In Mixer, all self-attention layers in the Transformer architecture are replaced by MLP layers, each of which conducts two mixing operations on the channel (embedding) and the patch (sequence) dimension respectively:
\begin{align}
    X'_{*: t} &:= W^{p1} \left( W^{p1} \cdot X_{*: t} \right) & \mathrm{(token\ mixing)} \label{eq:token-mix} \\ 
    Z_{*: t} &:= \left( X'_{t:*} \cdot W^{c1} \right) W^{c2} & \mathrm{(channel\ mixing)} \label{eq:chan-mix} 
\end{align}

Combining Eq. \ref{eq:token-mix} and \ref{eq:chan-mix}, we have:
\begin{align}
    Z_{L \times D} :&= (W^{p2} W^{p1} \textcolor{blue}{X})\cdot W^{c1} W^{c2} \nonumber\\
                    &= (W^{p2} W^{p1}) X (W^{c1} W^{c2}) \nonumber\\
                    &= W^{p}\ X\ W^{c} \nonumber\\
                    &= \Mix_\textrm{mlp}(X\mid \theta_\textrm{mlp}) \\
                    &= \textcolor{blue}{\widetilde{X}_\mathrm{mlp}} \nonumber,
\end{align}
where $\theta_\textrm{mlp} = \{W^{p}_{L \times L}, W^{c}_{D \times D}\}$, which are the weights \textit{learned} with GELU during token and channel mixing respectively.
Although $\Mix_\textrm{mlp}$ is not static, it is \textit{independent} of the input sequence since $\textcolor{blue}{X}$ does not enter $\theta_\textrm{mlp}$.

\section{Replacing Attention with Fourier Transform} \label{sec:fourier}

Similar to MLP-Mixer, \citet{lee2021fnet} proposed a new mixing scheme that replaces the \textit{learned, expensive} full-attention by a series of \textit{fixed, efficient} Discrete Fourier Transforms (DFTs). 
For simplicity, common tricks like layer norms and skip connections are omitted.

\subsection{Weighting Sequence using Twiddle Factors}
All self-attention layers in the Transformer architecture are substituted by Fourier layers.
Each Fourier layer conducts a two DFT: first over the embedding dimension $D$ (Eq. \ref{eq:1st-ft}) and then over the sequence dimension $L$ of the input tokens (Eq. \ref{eq:2nd-ft}).

\begin{align}
    Z_{t, j} &:= \sum^{D-1}_{d=0} \exp\left\{\frac{2\pi i}{D} d \cdot j\right\} X_{t, d} & \mathrm{(first\ DFT)} \label{eq:1st-ft} \\
    \widetilde{X}_{t, j} &:= \Real\left\{\sum^{L-1}_{k=0} \exp\left\{\frac{2\pi i}{L} j \cdot k\right\} Z_{t, k}\right\} & \mathrm{(second\ DFT)} \label{eq:2nd-ft}
\end{align}

Both Eq. \ref{eq:1st-ft} and \ref{eq:2nd-ft} are in the form of weighted average similar to Eq. \ref{eq:conv-depth} and \ref{eq:attn-sum}.

\subsection{Mixing Tokens with Fourier Transform}

The weighted average forms of the two DFT can be rewritten using corresponding Vandermonde matrices for the roots of unity up to a normalization factor:
\begin{align*}
    F^h &:= \frac{1}{\sqrt{D}}
        \begin{bNiceMatrix}
            w^0 & w^0 & w^0 & \cdots & w^0 \\
            w^0 & w^1 & w^2 & \cdots & w^{(D-1)} \\
            w^0 & w^2 & w^4 & \cdots & w^{2(D-1)} \\
            \vdots &\vdots &\vdots &\ddots &\vdots \\
            w^0 & w^{(D-1)} & w^{2(D-1)} & \cdots & w^{(D-1)(D-1)}
        \end{bNiceMatrix} \\
    F^s &:=\ \frac{1}{\sqrt{L}}
        \begin{bNiceMatrix}
            w^0 & w^0 & w^0 & \cdots & w^0 \\
            w^0 & w^1 & w^2 & \cdots & w^{(L-1)} \\
            w^0 & w^2 & w^4 & \cdots & w^{2(L-1)} \\
            \vdots &\vdots &\vdots &\ddots &\vdots \\
            w^0 & w^{(L-1)} & w^{2(L-1)} & \cdots & w^{(L-1)(L-1)}
        \end{bNiceMatrix},
\end{align*}

where $w = \exp\{-2 \pi i\}$.

With $F^h$ and $F^s$, we simplify Eq. \ref{eq:1st-ft} and \ref{eq:2nd-ft} to:
\begin{align}
    X'_{t: *} &:= X_t \cdot F^h & \mathrm{(first\ DFT)} \label{eq:1st-ft-simple} \\ 
    Z_{*: t} &:= \Real\left\{F^s \cdot X'_{*: t}\right\} & \mathrm{(second\ DFT)} \label{label:2nd-ft-simple} 
\end{align}

By combining Eq. \ref{eq:1st-ft-simple} and \ref{label:2nd-ft-simple}, we have:
\begin{align}
    Z_{L \times D} :&= \Real\left\{ F^s\ \textcolor{blue}{X}\ F^h \right\}  \label{eq:ft-mix} \\
        &= \Mix_\textrm{FT}(X) \\
        &= \textcolor{blue}{\widetilde{X}_{\mathrm{FT}}}, \nonumber
\end{align}

From Eq. \ref{eq:ft-mix}, we have the following observations. First, DFT results in a token mixing $\textcolor{blue}{\widetilde{X}_{\mathrm{FT}}}$ in the same formulation as that of the attention mixing $\textcolor{blue}{\widetilde{X}_{\mathrm{attn}}}$ from Eq. \ref{eq:axw}. However, $\Mix_{\mathrm{FT}}$ is static (\textit{not} learned) and \textit{independent} of the input sequence (Eq. \ref{eq:kern-seq}). Secondly, by using Fast Fourier Transform (Cooley–Tukey algorithm \citep{cooley1965fft,frigo2005fftw3}), the computational cost is in $O(L\log L)$, with much smaller space complexity (since the symmetric Vandermonde matrices can be computed/stored efficiently), compared to that of the full-attention $O(L^2)$. Furthermore, the order in which the two DFTs are applied does not matter. Lastly, different from mixing schemes using MLP layers, stacking FT layers is analogous to switching between the ``time'' and frequency domain.

% \begin{enumerate}
%     \item DFT results in a token mixing $\textcolor{blue}{\widetilde{X}_{\mathrm{FT}}}$ in the same formulation as that of the attention mixing $\textcolor{blue}{\widetilde{X}_{\mathrm{attn}}}$ from Eq. \ref{eq:axw}.
%     \item However, $\Mix_{\mathrm{FT}}$ is static (\textit{not} learned) and \textit{independent} of the input sequence (Eq. \ref{eq:kern-seq}).
%     \item By using Fast Fourier Transform (Cooley–Tukey algorithm \citep{cooley1965fft,frigo2005fftw3}), the computational cost is in $O(L\log L)$, with much smaller space complexity (since the symmetric Vandermonde matrices can be computed/stored efficiently), compared to that of the full-attention $O(L^2)$. 
%     \item The order in which the two DFTs are applied does not matter.
%     \item Different from mixing schemes using MLP layers, stacking FT layers is analogous to switching between the ``time'' and frequency domain.
% \end{enumerate}

\section{Sequence Modeling with State Space Models} \label{sec:ssm}

A different branch of mixing scheme was proposed by \citet{gu2020hippo}, employing and augmenting traditional State Space Models (SSMs) from control theory.
Specifically, the authors employ linear time-invariant (LTI) SSM parameterized by a set of structured matrices to summarize history and memorize long-range dependencies.

\subsection{High-order Polynomial Projection} \label{subsec:hippo}

The starting point of this line of work is the idea of using orthogonal, high-order polynomial projection (HiPPO) to summarize a theoretically \textit{optimal} representation of \textit{all} the past tokens \citep{gu2020hippo}.
The HiPPO operator is a two-step transform: (1) it first takes a \textit{one-dimensional} input signal up to time $t$, projects it onto orthogonal basis polynomials of order $N$ with either uniform (Legendre) or exponentially decaying (Laguerre) weights, and (2) it then extracts the coefficients of the basis polynomials as the best representation of the past until the current point in time:
\begin{align}
    \texttt{hippo}(X_{L \times 1} |_{<t}) = \texttt{coef}\left\{ \texttt{proj}_{\perp}(X) \right\}\quad \xrightarrow{\textrm{dimension expansion}}\quad \bm{x} \in \R^{L \times N}, \label{eq:hippo}
\end{align}
where $\bm{x}$ is a matrix that contains \textit{all} the system states of the SSM prior to time $t$.
In other words, row $t$ of the state matrix, $\bm{x}(t)^\top \in \R^{1\times N}$, summarizes the history of the input sequence before time $t$.

By leveraging a specially structured state transition matrices $A$ and $B$, integrating the following ordinary differential equation (ODE) of the SSM demonstrates SoTA results in summarizing and, in turn, reconstructing the history of the input signal \citep{gu2020hippo}:

\begin{align}
    \bm{\dot{x}}(t) := A \bm{x}(t) + B u(t)  , \label{eq:ode-hippo}
\end{align}
where the system input $u(t)$ is a single token of the \textit{one-dimensional} signal, i.e., $X_t$. Note that $A$ and $B$ are fixed, constant matrices when the method was first proposed \cite{gu2020hippo}.
However, these matrices can be learned through backpropagation, although its performance gain is not significant \cite{gu2021lssl}.
This learning process also incurs large overhead, especially for modeling high-dimensional feature space, which is improved through various mathematically techniques in follow-up work \cite{gu2022s4,gu2022s4d}.

\subsection{Multidimensional Projection} \label{subsec:lssl}

HiPPO has major two limitations: (1) The input signal is restricted to one dimension, and (2) Matrices $A$ and $B$ are not learned.
\citet{gu2021lssl} developed Linear State-Space Layers (LSSLs) to address the two challenges.

To work with multidimensional input, LSSL applies HiPPO \textit{independently} on each embedding dimension $d$ of the input sequence and concatenates the $D$ series of outputs as the representation of SSM at all times \footnote{According to their implementation: \url{https://github.com/HazyResearch/state-spaces/blob/main/src/models/s4/lssl.py}.}:

\begin{align}
    \bm{\dot{x}}^d(t)_{N\times1} &:= A \bm{x}^d(t) + B u^d(t)_{1\times1} \label{eq:lssl} \\
    {y}^d(t)_{1\times1} &:= C \bm{x}^d(t) + D u^d(t)_{1\times1} \nonumber.
\end{align}

More generally, for a sequence of length $L$, for all $t \in L$, we have:
\begin{align*}
    \bm{\dot{x}}^d_{N\times L} &:= A \bm{x}^d + B (\vec{u}^{\ d})^\top_{1\times L} \\
    (\vec{y}^{\ d})^\top_{1\times L} &:= C \bm{x}^d + D(\vec{u}^{\ d})^\top_{1\times L} ,
\end{align*}
where the $(\vec{u}^{\ d})$ is one column of the input signal $X_{L \times D}$, and the matrices $A, B, C$ and $D$ are all \textit{learnable} via backpropagation (through time).

Therefore, at any point in time $t$, the matrix $\bm{{x}}(t) \in \R^{N \times D}$ contains all the internal system states of the SSM, i.e., the coefficients of the orthogonal polynomials. 
Further, the tensor $\bm{{x}} \in \R^{N \times D \times L}$ contains all the systems states for $t \in [0, L]$.
For completeness, the update step size $\Delta t$ is also \textit{learnable} for discretization using Bilinear transform (empirically more performant than Euler method), i.e., $\bm{x}^d(t) \xrightarrow{} \bm{x}^d(t+\Delta t)$.

Although the parameter matrices and the step size are learnable, the training process is \textit{computationally prohibitive}, in part due to the expensive dimension expansion, i.e., the orthogonal polynomials. 
Consequently, they are \textit{fixed} in practice \citep{gu2021lssl}.

To reduce system costs and make the training more efficient, special tricks and parameterization have been developed, e.g., S4 \citep{gu2022s4}, S4D \citep{gu2022s4d}.

\subsection{Convolutional View of SSM} \label{subsec:conv-ssm}

Any LTI dynamic system can be viewed as the convolution between the input signal and its impulse response function, and so does the SSM described by Eq. \ref{eq:lssl}.
For simplicity, we drop the dimension $d$ and the matrix $D$ ($\approx$ skip connection) in the following derivation.
(The recurrent representation of SMM is theoretically interesting but practically infeasible due to its sequential nature, so it is not considered here.)

\begin{align*}
    \bm{\dot{x}}(t) :&= A \bm{x}(t) + B u(t) \\
    \bm{\dot{x}}(t) - A \bm{x}(t) &= B u(t) \\
    e^{-tA}\bm{\dot{x}}(t) - A e^{-tA} \bm{x}(t) &= e^{-tA} B u(t) \\
    \frac{1}{dt} \left[e^{-tA}\bm{{x}}(t)\right] &= e^{-tA} B u(t) \\
    \int_0^t \frac{1}{d\tau } \left[e^{-\tau A}\bm{{x}}(\tau )\right]d\tau &= \int_0^t e^{-\tau A} B u(\tau ) d\tau \\
    e^{-tA}\bm{{x}}(t) - \bm{{x}}(0) &= \int_0^t e^{-\tau A} B u(\tau ) d\tau \\
    \bm{{x}}(t) &= e^{tA}\bm{{x}}(0) + e^{tA}\int_0^t e^{-\tau A} B u(\tau ) d\tau \\
                &= e^{tA}\bm{{x}}(0) + \int_0^t e^{t-\tau A} B u(\tau ) d\tau \\
                &= e^{tA}\bm{{x}}(0) + \int_0^t \underbrace{ e^{tA} B}_{\textrm{basis function}} u(t-\tau ) d\tau  \numberthis{} \label{eq:conv-ssm-comm}\\
                &= e^{tA}\bm{{x}}(0) + \int_0^t h(t) \cdot u(t-\tau ) d\tau \\
                &= e^{tA}\bm{{x}}(0) + (h \odot u)(t)   \numberthis{} \label{eq:conv-ssm-states},
\end{align*}
where Eq. \ref{eq:conv-ssm-comm} is from the commutativity of the convolution, and $h(t)$ is the unit impulse response function of the SMM.

Substituting Eq. \ref{eq:conv-ssm-states} into output equation yields:

\begin{align*}
    y(t) :&= C \texttt{coef}\left\{ \texttt{proj}_{\perp}(\textcolor{blue}{X_{*:d}})\right\}(t) \\
          &= C \bm{x}(t)\\
          &= C \left[e^{tA}\bm{{x}}(0) + (h \odot u)(t) \right] \\
          &= \Mix_\textrm{ssm}(X_{*:d}\mid \theta_\textrm{ssm}) \\
          &= \textcolor{blue}{\widetilde{X}_\mathrm{ssm}} , \label{eq:conv-ssm-out}
\end{align*}
where $\theta_\textrm{ssm} = \{A, B, C\}$.

Hence, the resulting representation of the input sequence is a linear combination of the impulse response function.

Observe that (1) $\Mix_\textrm{ssm}$ \textit{depends} on $\textcolor{blue}{X}$, which enters $\theta_\textrm{ssm}$ via $A$, and (2) $\theta_\textrm{ssm}$ can either be \textit{learned} or be \textit{fixed} as specially structured matrices (the performance difference is not significant while fixing $\theta_\textrm{ssm}$ is much more efficient).

\subsection{More Efficient Polynomial Projection}

As discussed in \S\ref{subsec:lssl}, one key bottleneck is the dimension expansion as SSM requires a tensor of $\bm{{x}} \in \R^{N \times D \times L}$ to represent the entire system state.
One idea could be to use Product Quantization \citep{jegou2010pq} by first partitioning the input sequence into subspaces and learning a smaller prototype (an approximation) within each subspace \textit{during the initial pass of the training set}.
Then, project entire subspaces onto orthogonal polynomial basis functions, instead of doing it for every column/dimension.
Such a method can be combined with (tree-based) Locality Sensitive Hashing, which together could result in zero \texttt{matmul} operations \textit{during inference time} \citep{blalock2021nomatmul}.

% \bigskip

% \section{Laplace Transform} \label{sec:laplace}

% \answerTODO{} (potentially) More efficient SSM computation using Laplace Transform (LT)---The LT of the system output $y(t)$ is equal to the product between the LT of input signal $u(t)$ and the LT of the unit impulse response $h(t)$.

\section{Mixing Tokens with Convolution} \label{sec:sgcov}

Recently, \citet{li2022sgconv} proposed a pure convolutional architecture in place of the full attention block, achieving both lower system costs (15--50\% faster) and the same/higher model quality.

It employs concatenated parameter sets with decaying weights and applies three convolutions using real-valued DFT (1) on the concatenated parameter sets, (2) on the input sequence, and (3) between the two (inverse transform).
In turn, \textit{it can be regarded as a learnable, weighted FNet}.
As a result, its time complexity is in $O(L \log L)$, same as FNet.

\subsection{Memory Cost of SGConv}

Although the authors briefly mentioned that the memory complexity is also in $O(L \log L)$, we show here that it is likely to be in $O(L + \log L)$ instead.

Given an input sequence $X_{L\times D}$ and kernel dimension $k$, the number of kernels $s$ is calculated as the following:
\begin{align}
    s := \left\lceil \log_2(L/k) \right\rceil+1. \nonumber
\end{align}

Consequently, we need $s$ sets of learnable parameters, each of which corresponds to one kernel:

\begin{align}
    W^{\mathcal{K}} := \left\{ W_{k \times D}^{(1)}, W_{k \times D}^{(2)}, \dots, W_{k \times D}^{(s)} \right\}. \nonumber
\end{align}

$W^{\mathcal{K}}$ requires $\mathrm{{\Theta}}(s\times k \times D) \in O(\log L)$ space in total.

With the above parameters, SGConv instantiates a kernel $K$ concatenated from $s$ sub-kernels for \textit{every forward pass}:

\begin{align}
    K^{(i)} &:= \texttt{Interpolate}\left\{ W^{(i)} \right\} \in \R^{\left(k \cdot 2^{(i-1)}\right) \times D} \nonumber\\ 
    K &:= \texttt{concat}\left\{ K^{(1)},K^{(2)}, \dots, K^{(s)} \right\} \in \R^{L' \times D}, \nonumber
\end{align}
where $L' = \sum\limits^{s}_{i=1} k \cdot 2^{(i-1)} + k = L + \epsilon \approx L$.

Hence, the kernel $K$ takes $O(L)$ space. 
Since one layer of SGConv needs one set of parameters $W^{\mathcal{K}}$ and a concatenated kernel $K$, it requires $O(L + \log L)$ memory per layer.

This space complexity is smaller than that of S4 ($O(L + N)$ where $N=256$) \citep{gu2022s4}, and commensurate with existing attention approximations, e.g., Reformer ($O(L + \log L)$) \citep{kitaev2020reformer} and  Performer ($O(L)$) \citep{choromanski2020performer}.

\section{Impact of Context Length}
    \begin{figure}[t]
        \centering
        \protect\marginnote{Note that the y-axes do not start from zero.}
        \begin{adjustbox}{width=0.9\linewidth,center=0pt}
          \includegraphics[width=\linewidth]{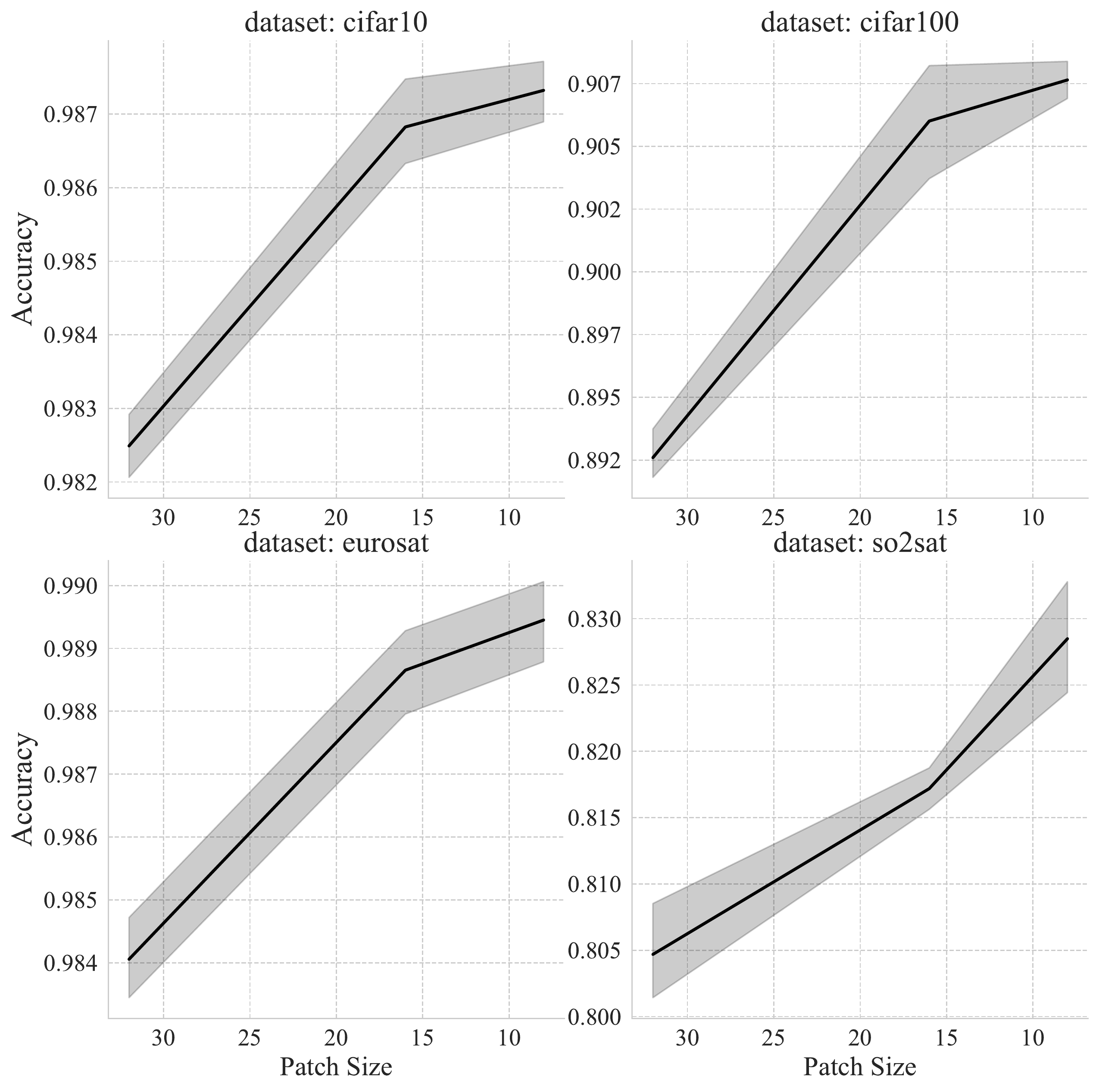}
        \end{adjustbox}
        \caption{Performance of pretrained ViT model on four datasets when varying the patch size of the image. The smaller the patch size, the longer the context length the model gets access to in each batch.}
        \label{fig:vit}
    \end{figure}
    % \marginpar[left text]{right text}

    \begin{figure}[t]
        \centering
        \begin{adjustbox}{width=1.1\linewidth,center=0pt}
        \begin{subfigure}{.50\linewidth}
          \centering
          \includegraphics[width=\linewidth]{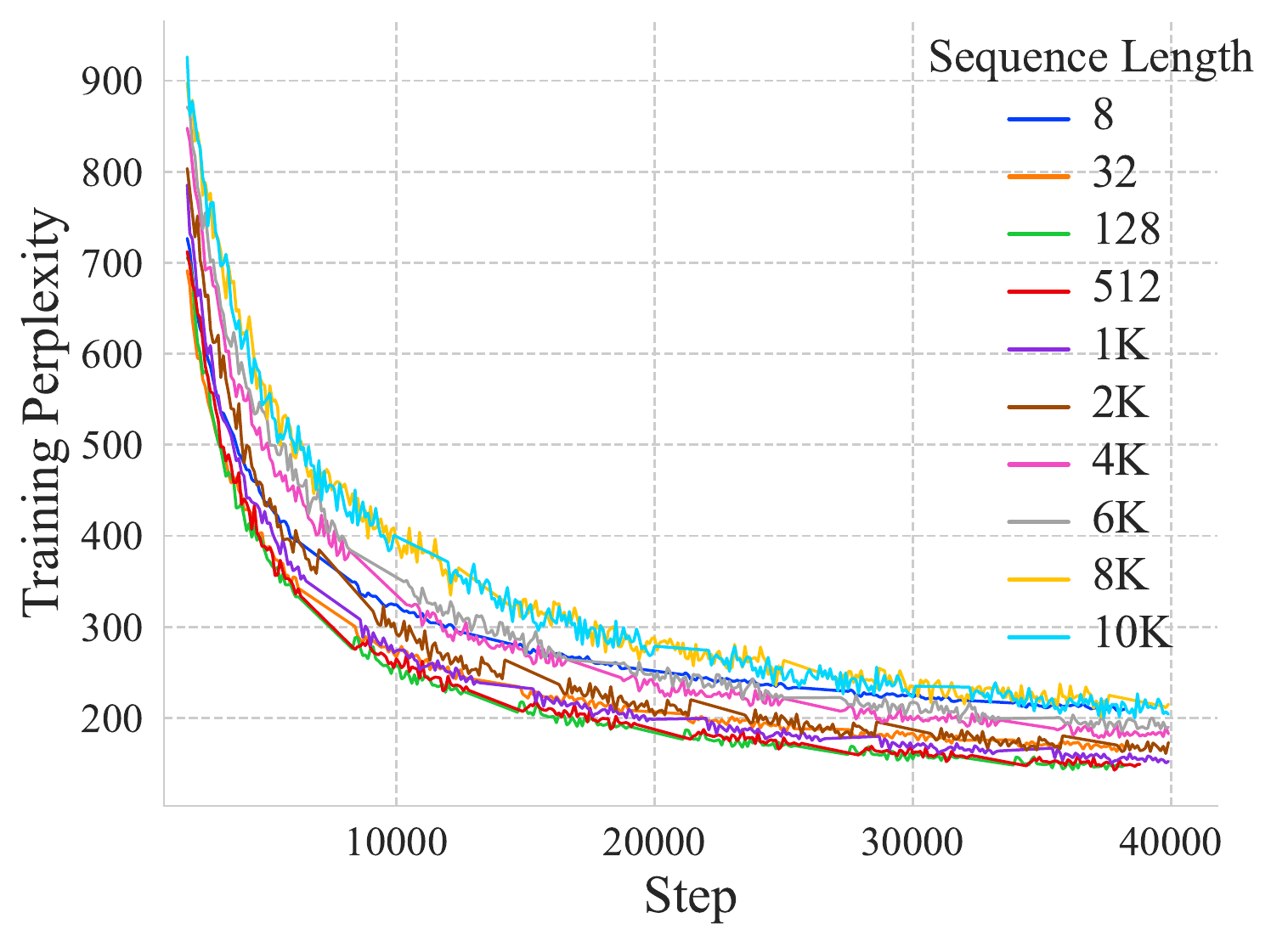}
          \caption{}
          \label{fig:train_ppl}
        \end{subfigure}%
        \begin{subfigure}{.50\linewidth}
          \centering
          \includegraphics[width=\linewidth]{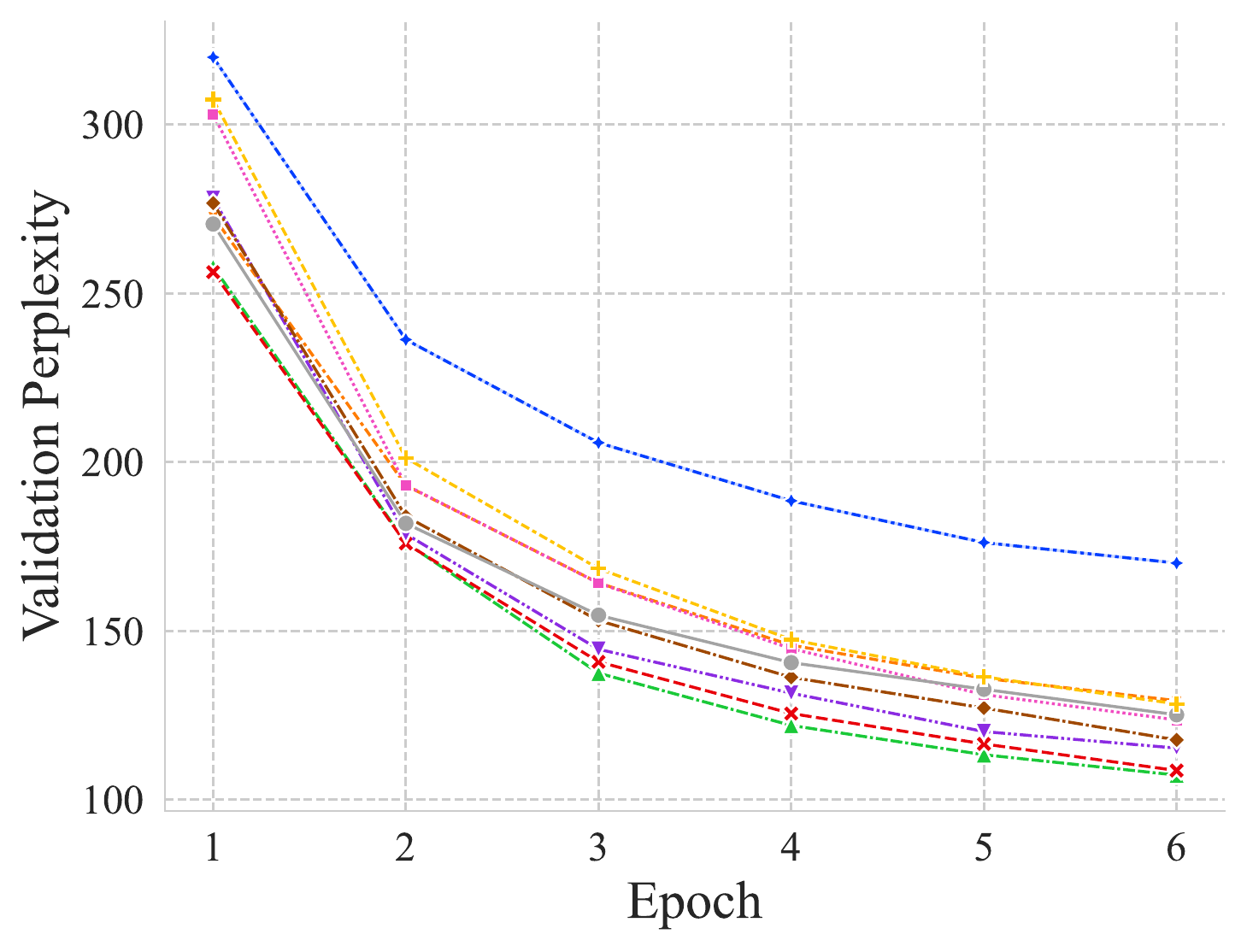}
          \caption{}
          \label{fig:val_ppl}
        \end{subfigure}
        \end{adjustbox}
        \caption{Training (a) and validation (b) loss of the vanilla Transformer model on WiKiText-103 with different context lengths.}
        \label{fig:ppl}
    \end{figure} 
    
    \begin{figure}[t]
        \centering
        \begin{adjustbox}{width=0.5\linewidth,center=0pt}
          \includegraphics[width=\linewidth]{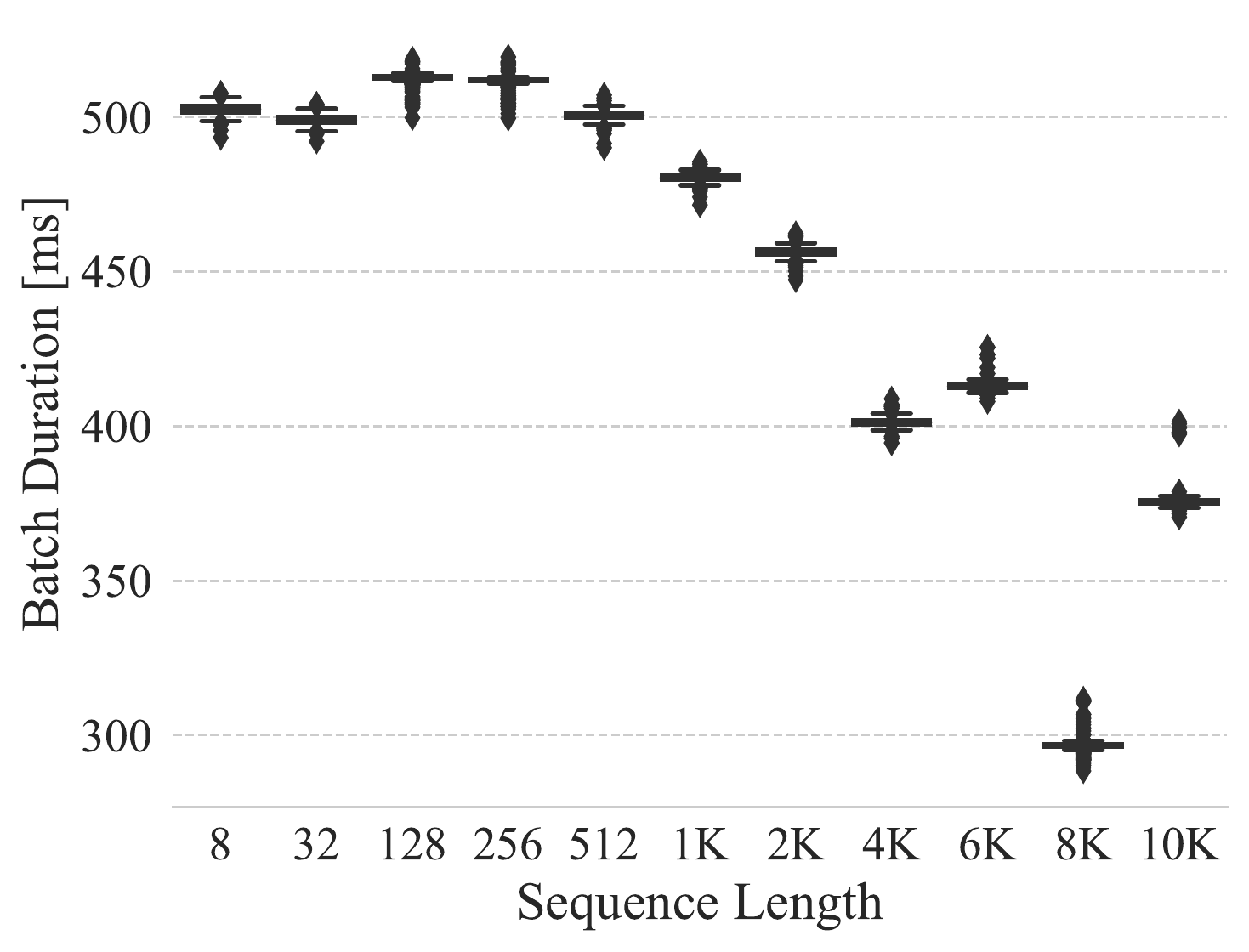}
        \end{adjustbox}
        \caption{Per-batch training time of the vanilla Transformer model on WiKiText-103.}
        \label{fig:duration}
    \end{figure}

In this section, we conduct experiments to investigate the impact of varying context lengths in sequence modeling.
One of the applications of long-context learning is on high-resolution images, e.g., fMRI, and satellite images.
Therefore, we first study the behaviors of Vision Transformers (ViT) \cite{dosovitskiy2020vit} when varying the sequence length.
To this end, we pretrain a base ViT on ImageNet (21k+) and fine-tune it on four datasets: CIFAR 10, CIFAR 100, EuroSAT \cite{helber2019eurosat}, and So2Sat \cite{so2sat}. 
The latter two are satellite datasets of higher resolution.
The downstream task for all four datasets is classification.
ViT partitions an image into patches and treats each patch as a token.
Therefore, the smaller the patches are, the longer the context the model gets access to in each batch.
Note that, although the dependencies between patches are preserved among patches, the structural information within a patch is destroyed as it is serialized during embedding.
We thus expect better performance when using smaller patch sizes since more structural information is kept.
Also, this performance increase is expected to be task-dependent, for example, predicting a dog needs far fewer details than identifying the type of vehicle in a satellite image.
We repeat the experiments for six runs on the ETH Euler cluster with two GeForce RTX 3090 GPUs.
As expected, model quality does increase as the patch size gets smaller (Fig. \ref{fig:vit}), and the magnitudes of such increase differ by task.
However, we notice that performance improvement is not significant in this experiment, and the trends tend to plateau at the end. 
Therefore further investigations with different tasks and more find-grained increments of context length are needed.

To avoid being constrained by the fixed architecture of pretrained models, we write a vanilla Transformer for the following experiments so that we can vary the sequence length at will.
The Transformer model consists of three layers of encoder blocks and one MLP layer as the prediction head, the decoder.
We implement the positional embedding described in the original work \cite{vaswani2017attention} and keep it autoregressive by implementing attention masks.
The model is trained on the WikiText-103 dataset from scratch using one NVIDIA A100 GPU.
The dataset contains 103M tokens and has a vocabulary size of 98K.
The downstream task is language modeling, and we use perplexity as the performance metric.
This time we vary sequence length from 8 to 10K with log2 step sizes.
Similar to experiments conducted by \citet{wu2022memorizing}, we keep the number of tokens per batch constant ($2^{14}$) while sweeping the sequence length.
To achieve this, we dynamically adjust the batch size length given the sequence length.
By doing so, \textbf{the model gets access to different context lengths while still seeing the same total number of tokens in each batch}.

The average length of the articles in the WiKiText-103 dataset is 3.6K words \cite{bai2021segatron}, which is expected to be the ideal context length for this task.
However, the model reaches its peak performance with a sequence length around 128-512 (Fig. \ref{fig:ppl}), because this sequence length has reached the capacity of traditional Transformer models (e.g., BERT \cite{devlin2018bert}, GPT-2 \cite{radford2019gpt2}).
In addition, we observe that the training time \textit{decreases} with the sequence length (Fig. \ref{fig:duration}).
Since we keep the total number of tokens per batch constant, this result illustrates that doing multiple small attention passes on same number of tokens is more expensive than doing one larger pass to compute (bigger) attention matrices over more tokens.
This tradeoff forms a constrained optimization problem with the performance being the objective and runtime being the constraint.

\section{Architecture for Million-Scale Dependencies}

\begin{table}[t]
\centering
\marginnote{This summary table is not exhaustive.}
\begin{tabular}{rl} 
    \toprule
    \textbf{Max. Context Length} & \textbf{Solution Category}                               \\
    \midrule
    512--2K              & Full attention (e.g.,\cite{devlin2018bert,vaswani2017attention})                                \\
    $\sim$65K           & Approximated attention (e.g., \cite{choromanski2020performer,wang2020linformer,beltagy2020longformer}); Memory I/O optimization \cite{dao2022flashattention}    \\
    264K                & Neural Turing Machine with multi-head attention \cite{wu2022memorizing} \\
    $\sim$1M*            & Token mixing (e.g., \cite{tolstikhin2021mlpmixer, lee2021fnet, li2022sgconv}); State Space Models (e.g., \cite{gu2020hippo,gu2022s4,gu2022s4d}) \\
    \bottomrule
\end{tabular}
\vspace{2mm}
\caption{Categorization and comparison between existing solutions in terms of maximum context length. *The 1M context length is achieved on toy examples (e.g., memorizing random sequences), but not with real-world downstream tasks.}
\label{tab:length}
\end{table}

In previous sections, we demonstrated that existing long-context models can be viewed as various token-mixing schemes. 
These schemes strive to make every token available to any other tokens within the context window and then, compute the weighted average of the context (together with the original embedding) to achieve inductive bias.
We have also demonstrated the tradeoff between performance and efficiency when varying the context length.
Table \ref{tab:length} summarizes and compares on a macro level the existing solutions to long-context learning problem, in terms of the maximum context length each model can handle. 
Note that it demonstrates only the \textit{feasibility} of working with such context length (i.e., constrained by the model architecture, and in turn, system resources), but not the model \textit{quality}.
In general, as the context get larger, model performance first increases (due to the access to more context) and then drops quickly as the length exceeds the model capacity \cite{bai2021segatron, wu2022memorizing, gu2022s4, gu2020hippo}.
Thus, although the optimization landscape seems to be near-exhausted, million-scale context still appears to be the pinch point thereof.

\subsection{Huge Sparse Models with Conditional Computation}
    
Sparse models using conditional computation \cite{bengio2015conditional} have (re)emerged as a promising direction towards huge model capacity, while keeping efficiency costs at bay.
The model that has attracted the most attention recently is the sparsely-gated Mixture of Experts (MoE) \cite{shazeer2017sparsemoe}.
It has shown to be able to scale to billions of parameters, while incurring a fraction of the costs of utilizing their full capacity (e.g., \cite{fedus2021switch,chowdhery2022palm,zhou2022expertchoice}).
Such results are due to the fact that these models are only partially activated at any time.
This partial activation brings about one crucial benefit, namely, \textcolor{blue}{model specialization} --- different parts of the model specialize at different tasks.
It has been shown that different experts in the sparse MoE model are highly specialized in terms of syntax and/or semantics in NLP tasks \cite{shazeer2017sparsemoe}. 

Furthermore, the component that is responsible for sparsely activating the model is the ``router'' that selects expert(s) and forwards tokens to them. In turn, this routing operation will activate only part of the model represented by the expert(s).
This router is typically made of MLP layers and learns to which expert(s) to forward which tokens.
Most importantly, it has demonstrated a filtering/pruning effect --- dropping redundant tokens while maintaining model performance \cite{fedus2021switch,zhou2022expertchoice}.
This feature allows the model to focus only on the most relevant part of an example (e.g., the delineation of a dog in an image) and ignore the relatively unimportant parts (e.g., the background and other objects in the image), which we call {\textcolor{blue}{concentrated learning}}.

\subsection{Learning Million-Scale Dependencies using Sparse Models}
The key tradeoff when dealing with long-range dependencies lies between resource efficiency and the amount of long/short-term information kept in memory, which dictates the learning paradigm and model capacity.
In the ideal case, model would store all history (and the future dependencies if not autoregressive) and make decisions accordingly.
For million-scale dependencies, such an ideal scenario is not realistic for two reasons: (1) it would induce a complexity \textit{at least} as large as the amount of information stored in terms of both space and compute, and (2) the memorized information becomes increasingly stale as the learning process goes since what is stored is the latent space rather than the raw tokens \cite{wu2022memorizing}.

\begin{figure}[t]
        \centering
        \begin{adjustbox}{width=1.2\linewidth,center=0pt}
          \includegraphics[width=\linewidth]{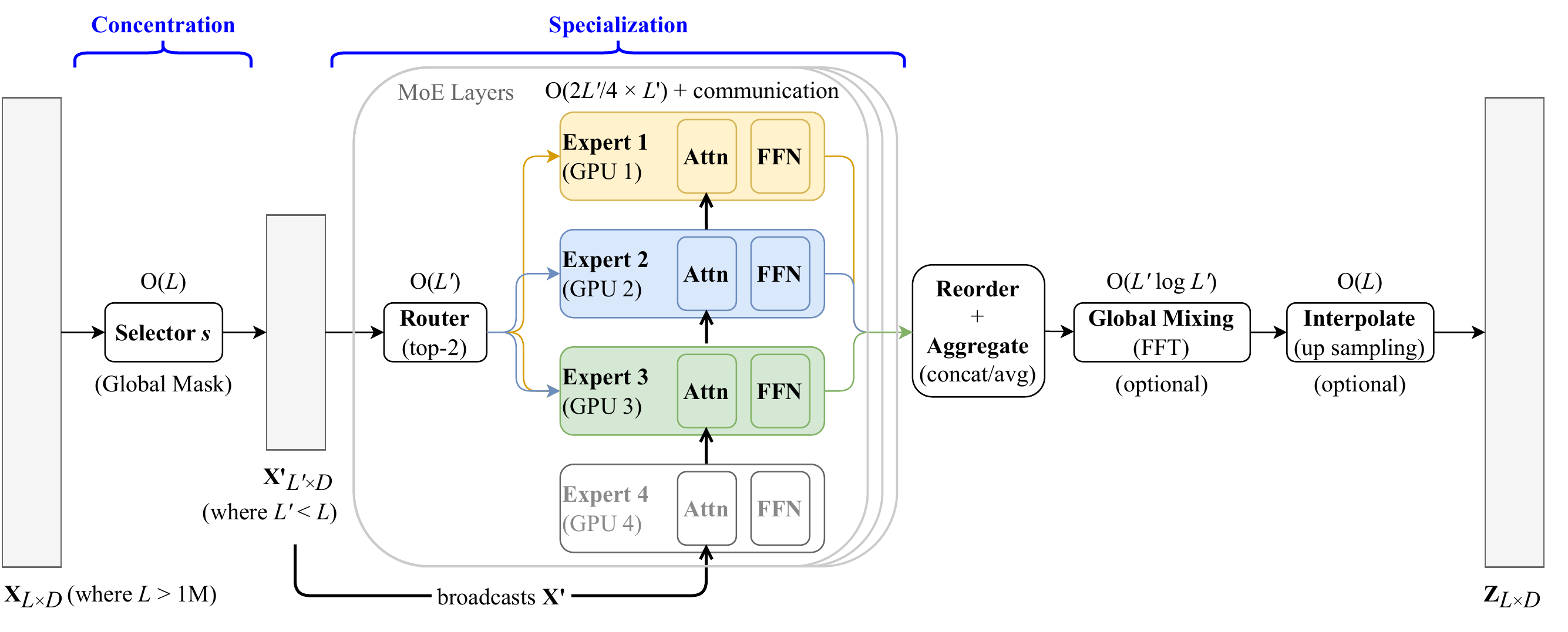}
        \end{adjustbox}
        \caption{System architecture of the proposed sparse model for million-scale dependencies.}
        \label{fig:arch}
    \end{figure}

Although sparse MoE models are not specifically designed for modeling long-range dependencies, we believe they are a good fit for the following reasons.
First, \textit{specialization} allows for increasing model capacity while keeping efficiency cost (training/inference time) relatively constant.
Second, \textit{concentration} makes the attention mechanism selective, i.e., only memorizing relevant information as opposed to all historical (and future) data.
Therefore, we propose an architecture based on sparse MoE (Fig. \ref{fig:arch}).

Inspired by the routing mechanism in sparse MoE, we introduce the \texttt{Selector} to filtering tokens \textit{before} computing attention.
This process aims to help the model concentrate on relevant parts of the samples, which also reduces the resource usage from the get-go.
Then, the selected tokens are passed through a distributed MoE layers, within which each expert resides on one device and handles part of the sequence.
Such a partitioning allows experts to compute \textit{full-attention} matrices for the tokens they receive and specialize at specific parts of the task.
Similar to the \texttt{Router}s in the MoE layers, the \texttt{Selector} can also be trained with the help of auxiliary losses that is added to the training loss during backpropagation.
The processed tokens are reordered and aggregated after being processed by the MoE layers. 
Additionally, a global mixing by means of fast Fourier Transform for wider context accessibility and an up-sampling processing through interpolation (\textit{k}NN or linear) to restore the dimensionality of the embedding could be applied. 
The theoretical complexity is asymptotically linear (and worst-case log linear) to the sequence length.

\subsection{Objective Functions}

Although the \texttt{Selector} can be trained with auxiliary losses similar to the \texttt{Router}s \cite{shazeer2017sparsemoe,chowdhery2022palm,fedus2021switch} in the MoE blocks, it is not the preferred option since these losses are often intuition-based and generally hard to evaluate their effectiveness. 
Instead, we aim to train the model end-to-end, without singling out the \texttt{Selector}.
Additionally, since the main goal of the \texttt{Selector} is to learn to prone/filter out unimportant tokens (e.g., the background, other objects in the image), a threshold parameter as a ``nob'' is useful for controlling the dropout rate, that is:
\begin{equation}
    s_{\psi, \tau}: X_{L \times D} \mapsto X'_{L'\times D}, \label{eq:selector}
\end{equation}
where $\tau$ is the pruning threshold, and $L \gg L'$.

With the selector defined, the objective reads:
\begin{equation}
    \max_{\theta, \psi}\pr \left( \vec{y} \mid Z = \frac{1}{M} \sum^{M}_{i=1} g_i \cdot f_\theta\left( s_{\psi, \tau}(X) \right) \right), \label{eq:obj} 
\end{equation}
where $M$ is the number of ensembled experts, and $g_i$ is the gating weight.

Alternatively, the objective (Eq. \ref{eq:obj}) can be in an adversarial form.
\citet{miladinovic2022gandropout} demonstrated the potential of learning a dropout model via a GAN-like formulation.
We modify the objective in an analogous way by creating a max-min game between the \texttt{Selector} and the predictor:
\begin{equation}
    \max_{\theta}\min_{\psi}\pr \left( \vec{y} \mid Z = \frac{1}{M} \sum^{M}_{i=1} g_i \cdot f_\theta\left( s_{\psi, \tau}(X) \right) \right). \label{eq:adverse} 
\end{equation}

Specifically, the \texttt{Selector} (adversary) tries to drop as many informative tokens as possible, whereas the predictor still aims for higher likelihood.
However, our preliminary experiments show that this adversarial objective often yields a too powerful \texttt{Selector} that the predictive part needs to be retrained with $s$ being frozen.

\subsection{Distributed Attention}

    \begin{figure}[t]
        \centering
        \begin{adjustbox}{width=1.5\linewidth,center=0pt}
          \includegraphics[width=\linewidth]{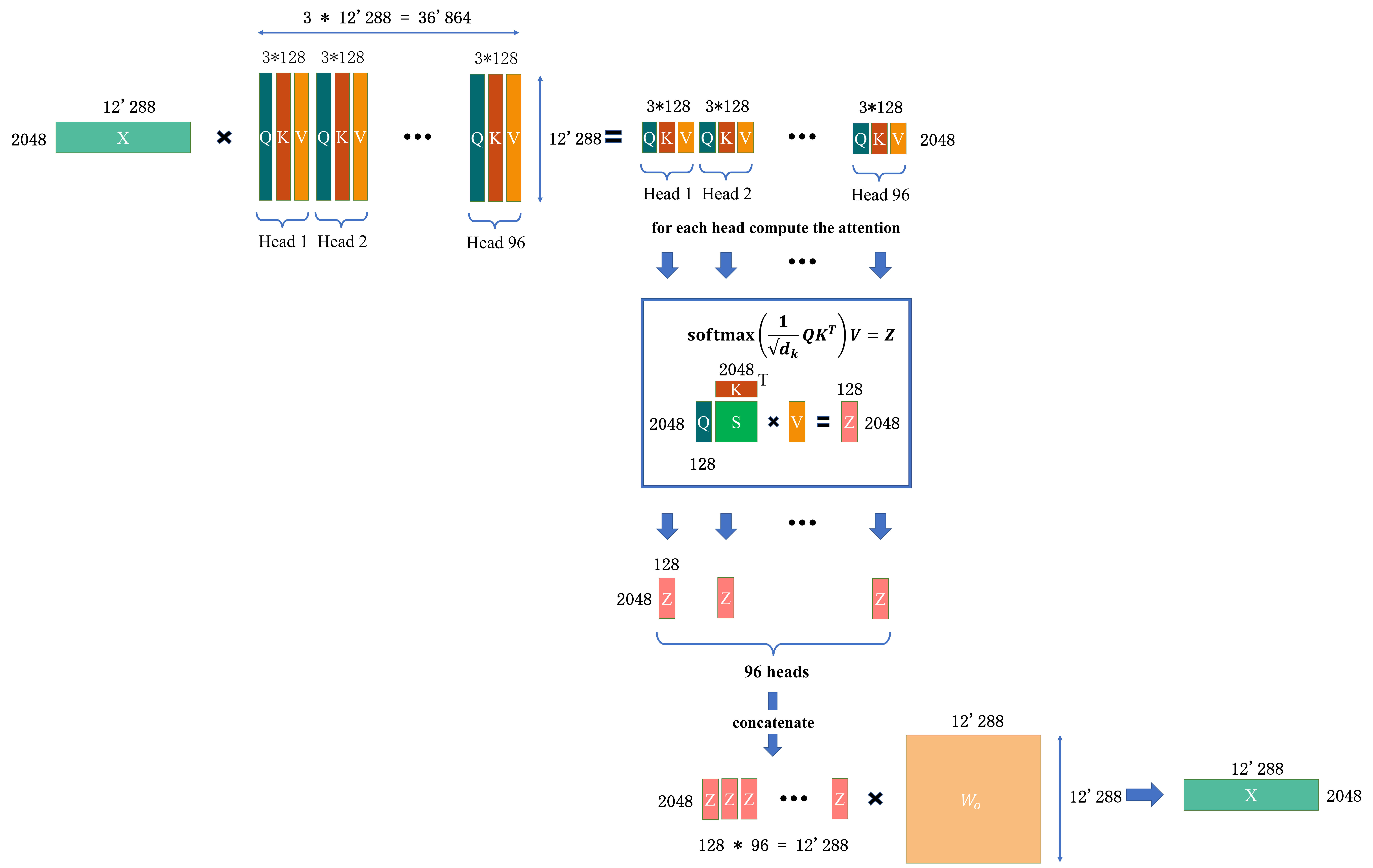}
        \end{adjustbox}
        \caption{Standard multi-head attention with all sizes taken from OPT-175B \cite{zhang2022opt} and GPT-3 175B \cite{brown2020gpt3} models.}
        \label{fig:standard_attn}
    \end{figure}

\begin{figure}[t]
        \centering
        \begin{adjustbox}{width=1.1\linewidth,center=0pt}
          \includegraphics[width=\linewidth]{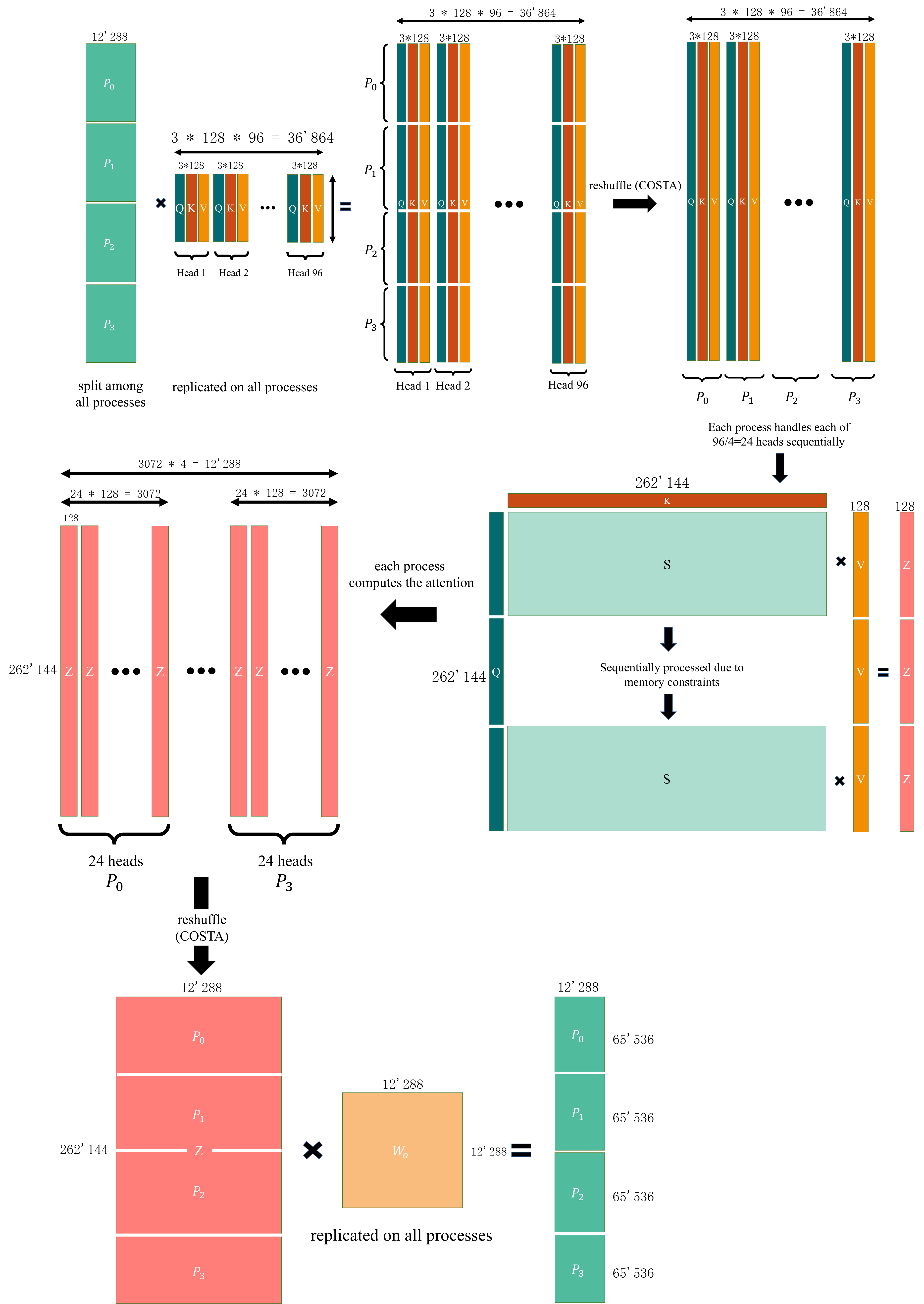}
        \end{adjustbox}
        \caption{Distributed multi-head attention with the sequence size being scaled, and all the other sizes are taken from OPT-175B \cite{zhang2022opt} and GPT-3 175B \cite{brown2020gpt3} models.}
        \label{fig:distributed_attn}
    \end{figure}

One critical element of the proposed architecture (Fig. \ref{fig:arch}) is the distributed computation of the attention matrix.
In standard attention computation, the entire matrix is produced on a single device (Fig. \ref{fig:standard_attn}).
However, for million-scale context, the sequence dimension is so large that the input matrix $X$ cannot fit in a single device memory and thus has to be distributed. 
For example, in Fig. \ref{fig:distributed_attn}, we assume the sequence dimension corresponds to the number of pixels in a $512 \text{px} \cdot 512 \text{px} =$ 262'144px image.
Taking all other dimensions from OPT-175B \cite{zhang2022opt} and GPT-3 175B \cite{brown2020gpt3} models, $X$ has to be distributed among $N$ GPUs within the same node (in this case, $N=4$).

The distributed algorithm works as follows. Firstly, we split the design matrix $X$ along the sequence dimension into $N$ partitions $\{P_i\}^N$.
Then, we replicate the parameter matrix along the unchanged dimension on \textit{all} devices.
Each device computes the attention matrix for all attention heads but only with the samples in their own partitions.
Next, we use COSTA \cite{kabic2021costa} to efficiently shuffle partitions so that each device has the attention scores for \textit{all} samples but with only $1/N$ heads.
This step prepares for the \texttt{softmax} calculation that requires all examples for each embedding dimension, and it can be computed \textit{sequentially} by each device.
Lastly, we use COSTA to reshuffle the data for the second time to compute the final linear projection with $W_0$ replicated on each device.
We implement this algorithm with NCCL MPI API from NVIDIA in Python.

\begin{figure}[t]
        \centering
        \begin{adjustbox}{width=0.75\linewidth,center=0pt}
          \includegraphics[width=\linewidth]{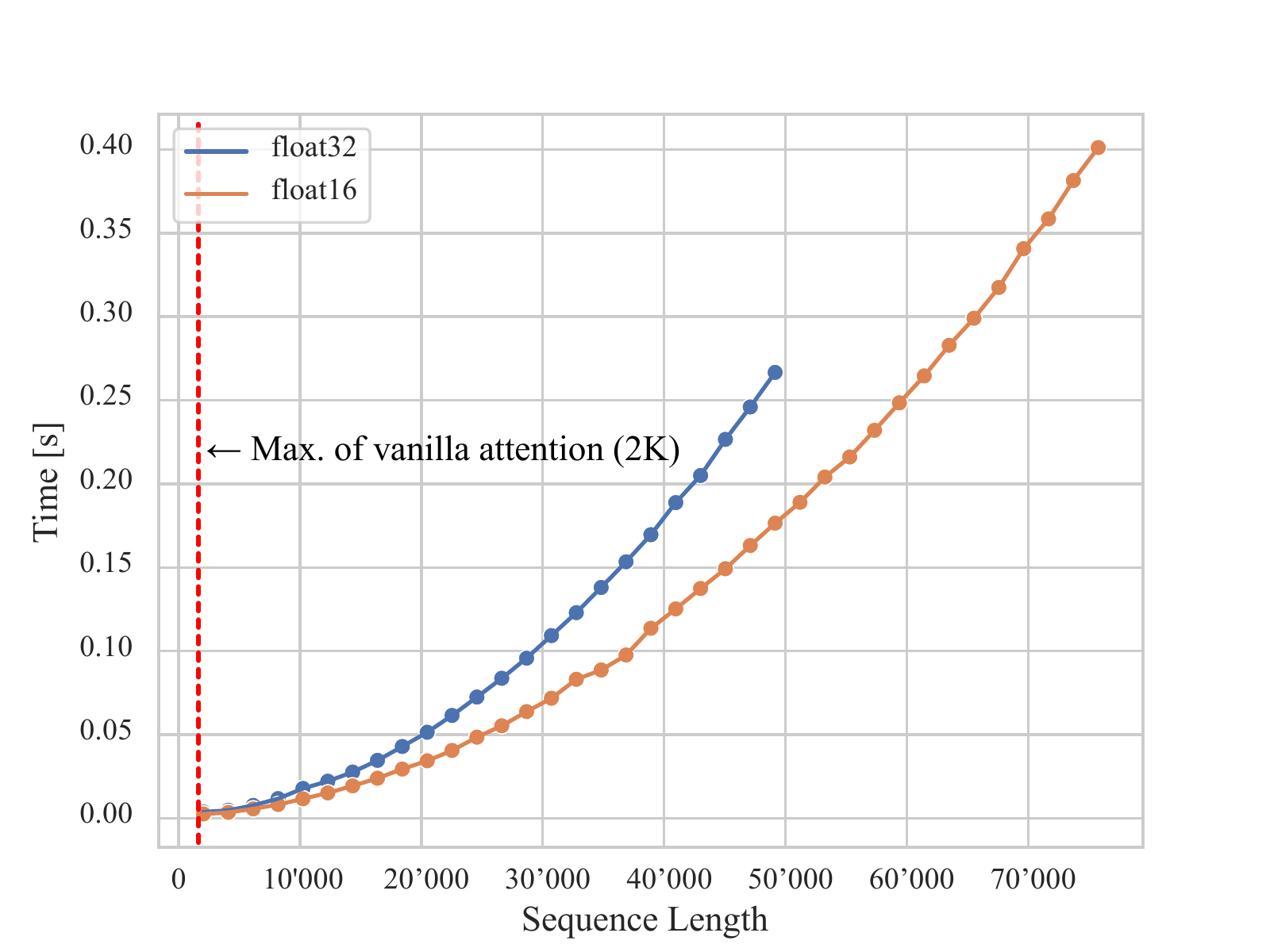}
        \end{adjustbox}
        \caption{Forward pass performance of the distributed attention algorithm on four GeForce RTX 3090 GPUs. Note that, without distributing the computation, the maximum sequence length that the vanilla attention implementation can handle is around 2K.}
        \label{fig:feasibility}
    \end{figure}

% \enlargethispage{\baselineskip}
To test the feasibility of this algorithm, we scale the length of the input sequence of the attention computation on four GeForce RTX 3090 GPUs (Fig. \ref{fig:feasibility}).
As a result, with four GPUs, the algorithm can scale the attention computation to a sequence length of almost 80K, which is 40$\times$ the maximum length a vanilla attention implementation can handle, with a near-linear growth in time complexity.

\section{Conclusion}

In this work, we first categorize and compare existing solutions to long-sequence modeling. 
By formulating them in a unified template mathematically, we pinpoint the nature shared among most prior works: making both global and local context available when computing attention scores through various token mixing schemes.
Next, we highlight the tradeoff between resource efficiency and the amount of memorized long/short-term history in such mixing schemes.
To model million-scale dependencies while keeping resource usage at bay, we then propose a distributed learning system inspired by recently proposed sparse MoE models of huge capacity, aiming to exploit two main features thereof: model specialization and concentrated learning.
As the first step towards building this system, we propose a distributed algorithm for computing attention matrices for million-scale sequences.
We demonstrate in experiment that our algorithm can scale the attention computation by almost 40$\times$ in terms of maximum sequence length with a near-linear time complexity, compared to that of vanilla attention implementation.
Although there is still much work to be done, we believe that this work is an instrumental step towards modeling million-scale dependencies.

\clearpage
% \medskip
{\footnotesize \bibliographystyle{abbrvnat}
\bibliography{references}}

\end{document}